# Information Retrieval Systems Adapted to the Biomedical Domain[*]

Mónica Marrero[1], Sonia Sánchez-Cuadrado[1], Julián Urbano[1],
Jorge Morato[1] and José-Antonio Moreiro[2]

[1]Department of Computer Science. University Carlos III of Madrid, Spain
mmarrero@inf.uc3m.es    ssanchec@ie.inf.uc3m.es    jurbano@inf.uc3m.es    jmorato@inf.uc3m.es

[2]Department of Library Science. University Carlos III of Madrid, Spain
jamore@bib.uc3m.es

**Abstract:** The terminology used in Biomedicine shows lexical peculiarities that have required the elaboration of terminological resources and information retrieval systems with specific functionalities. The main characteristics are the high rates of synonymy and homonymy, due to phenomena such as the proliferation of polysemic acronyms and their interaction with common language. Information retrieval systems in the biomedical domain use techniques oriented to the treatment of these lexical peculiarities. In this paper we review some of the techniques used in this domain, such as the application of Natural Language Processing (BioNLP), the incorporation of lexical-semantic resources, and the application of Named Entity Recognition (BioNER). Finally, we present the evaluation methods adopted to assess the suitability of these techniques for retrieving biomedical resources.

## 1. Introduction and Characteristics of the Biomedical Domain

The increase of documentation and the urge to locate relevant answers turns biomedical literature into an excellent candidate for the application of textual treatment techniques for Information Retrieval (IR). Retrieval systems in Biomedicine use controlled vocabularies (e.g. MeSH, INSPEC Thesaurus, Gene & Plant Ontology) in order to improve searches, following G. Salton's proposals in the sixties for general purpose IR systems. Techniques such as Natural Language Processing (NLP), Information Extraction (IE) and data mining have become essential to process, identify and infer information from this enormous amount of data. These techniques, usually referred to as text mining in the biomedical literature, are used on IR systems (Table 1), and they adopt certain peculiarities when applied in this area, due to the own characteristics of the domain and its terminology.

Biomedical research is characterized for being divided into highly specialized sub-areas conceptualized from different points of view, which tends to narrow the perspective, impeding the establishment of connections between discoveries (Weeber et.al., 2000) and cross information retrieval. The terminology in the biomedical domain is characterized by the constant growth on the number of terms for a concept in different areas and different registers, as well as the proliferation of new acronyms that bear high polysemy and lexical synonymy. New terminology lacks of patterns or constant regular expressions to allow their automatic identification with rules on the use of capital letters, sequences of special characters, alphanumeric sequences, etc. Indeed, one third of the occurrences of such terms are variants of others (orthographic, permutation, insertion or deletion, e.g. FOXP2 and FOXP3, MRP2 and MRP3, etc.) (Jacquemin, 2001).

A concept in this domain may have six or seven synonyms, because they show up in different areas, due to commercial issues, lack of consensus between experts or because of their scientific evolution or obsolescence (Table 2). For example, pharmaceutical products like paracetamol are also known by their ICD name (International Statistical Classification of Diseases and Related Health Problems), acetaminophen. The synonym in the IUPAC naming system (International Union of Pure and Applied Chemistry) is N-(4-hydroxyphenyl)ethanamide, which is equivalent to the chemical formula $C_8H_9NO_2$, and which is also known

---

by the code NO2 BE01 of the Anatomical Therapeutic Chemical classification system (ATC) of the World Health Organization. Its commercial name also varies across countries. In the USA it is known as Tylenol or Datril, in the UK it is called Tylex CD or Panadeine, in Spain it is referred to as Panadol, Termalgin, Efferalgán, Gelocatil or Apiretal, and in Mexico it is usually known as Tempra.

Table 1. Freely available biomedical search engines

| Tool | Corpus | Type | Main techniques |
|---|---|---|---|
| **NovoSeek** www.novoseek.com | MEDLINE, US Grants, Others | Commercial: Bioalma | Retrieval with self-constructed dictionaries |
| **PubFocus** www.pubfocus.com | PubMed, MEDLINE | Research | Dictionary-based retrieval with the NCI Thesaurus and MGD (Mouse Genome Database). Ranking using the Journal Citation Reports impact factor |
| **BioMed Search** www.biomedsearch.com | PubMed, MEDLINE | Commercial: BioMed Search | Retrieval with clustering |
| **XploreMed** www.ogic.ca/projects/xplormed/info | MEDLINE | Research | Use of NLP techniques (stop words, disambiguation and stemming with TreeTagger) |
| **Path Binder (prototype)** pathbinderh.plantgenomics.iastate.edu | PubMed, MEDLINE (partially) | Research | Dictionary-based retrieval with the Gene & Plant Ontology, Enzyme Nomenclature and MeSh. Includes taxonomy filter |
| **Textpresso** www.textpresso.org | C. Elegans literature. New papers can be added | Research | Retrieval with self-constructed ontologies of terms and processes |

On the other hand, polysemy affects both to common names (e.g. the English term "cold" may refer to the Chronic Obstructive Lung Disease, to something cold or to have a cold) and to specialized terminology in different fields of Biomedicine. For example, NF2 is the name of a gene, the protein that produces the gene and the disease produced by its mutation (Bodenreider, 2006), and NFKB2 denotes a family of two individual proteins, but belonging to different species: human and chicken. Acronyms augment the number of synonyms in the scientific literature, and it is estimated that 80% of them are ambiguous (Liu et.al., 2002). Indeed, once every five articles there appears a new acronym which is usually the same as many others previously coined (Spasic and Ananiadou, 2005).

Table 2. AIDS descriptors in MeSH

| 1979-1982 | Immunologic Deficiency Syndromes |
|---|---|
| 1984-1986 | Human T-Cell Leukemia Virus/HTLV/LAV |
| 1986-1992 | HIV |
| 1992- | HIV-1/HIV-2 |

## 2. Resources for the Representation of Knowledge in Biomedicine

In order to face the terminological problems, Knowledge Organization Systems (KOS) are developed, like gazetteers, classifiers, thesauri and ontologies (Table 3). However, these resources do not always simplify the process of indexing and retrieving information. For example, the Unified Medical Language System (UMLS) is used with MEDLINE to provide the possibility for query expansion with related terms. Query expansion with a controlled vocabulary improves effectiveness, especially when expanding with synonym terms (Hers et.al., 2000). But phenomena like poly-hierarchy lower the precision of the results. For example, in MeSH, the HIV might be found under any of these hierarchical branches: RNA Virus Infections (C02.782), Sexually Transmitted Diseases (C02.800), Slow Virus Diseases (C02.839) or Immune System Diseases (C20).

Moreover, in the case of Biomedicine the complexity of these resources is bigger than in other fields, because of the knowledge to be represented (entities, processes, functions, interactions, etc.) and the diversity of areas it belongs to at the same time. However, within Biology, the enthusiasm for ontologies has been accompanied by a general lack of awareness as to what exactly ontologies are and how to use them (Soldatova and King, 2005). The Open Biomedical Ontologies Foundry (OBO), started on 2001, works for redirecting this tendency, offering rules for the construction of ontologies in this domain while covering many sub-areas (anatomy, genomics, proteomics, experimental conditions, metabolomics, phenotype, etc.).

Table 3. Freely available Knowledge Organization Systems for Biomedicine

| Resource | Domain | Contents | Characteristics |
| --- | --- | --- | --- |
| **UML** Unified Medical Language System www.nlm.nih.gov/research/umls | Biomedical- and health-related concepts | 1000000+ terms | Uses an upper ontology to integrate diverse resources, such as SNOMED, MeSH and GO |
| **SNOMED-CT** Systematized Nomenclature of Medicine Clinical Terms snob.eggbird.eu | Medical records | 400000 terms | English, Spanish (free) and German (not free). Implemented with OWL |
| **GO** Gen Ontology www.geneontology.org | Genetic terms, grouped by molecular functions, biological processes and cellular locations | 9000+ terms | 50% of the terms from GO could be mapped to MeSH and SNOMED (McCray, 2002) |
| **MeSH** Medical Subject Headings www.nlm.nih.gov/mesh | Biomedicine, including nursing, veterinary and sanitary systems | 22995 descriptors | Used for automatic indexing and manual indexing of MEDLINE |

Other initiatives focus on the construction of upper-domain ontologies, such as the OBR framework (Ontology of Miomedical Reality), in order to integrate domain ontologies from anatomy, physiology and pathology (Rosse et.al., 2005). There are many other examples using upper ontologies. For instance, the Ontology for Molecular Biology (MBO), which comprises an upper ontology with concepts like "being", "abstract object", "event", etc. The Basic Formal Ontology (BFO) is an upper-domain ontology based on the Descriptive Ontology for Linguistic and Cognitive Engineering (DOLCE) and the Suggested Upper Merged Ontology (SUMO), both upper ontologies, and it provides support to domain ontologies development for scientific research inside the OBO Foundry. Also, BioTop and ChemTop (Stenzhorn et.al., 2008), two upper-domain ontologies, are based on BFO and the OBO Relation Ontology (RO). DOLCE is also the basis for an upper ontology called Simple Bio Upper Ontology (SBUO) (Rector et.al., 2006), and SUMO is the basis for the upper ontology developed in the BioCaster project to monitor infectious diseases in the Asiatic countries (Collier et.al., 2007), which integrates different knowledge sources in different languages. There are also meta-ontologies for the biomedical domain, such as Bio-Zen, which unifies different representation schema: DOLCE, Simple Knowledge Organisation System (SKOS), Semantically Interlinked Open Communities (SIOC), Friend Of A Friend (FOAF), Dublin Core and Creative Commons (Samwald and Adlassnig, 2008).

In line with the objectives of the Semantic Web, there have been developments in the last few years with works like Bio2Rdf (http://bio2rdf.org) and the Linking Open Drug Data Project (http://esw.w3.org/topic/HCLSIG/LODD). These projects offer the possibility of integrating schema about genes, proteins, drugs and clinical trials, both with other schema specific to Biomedicine and with generic ones (e.g. DBPedia http://dbpedia.org).

All these initiatives have an effect on the results of information retrieval and extraction systems. Given the complexity of the terminology in Biomedicine, it has been pointed out that the use of techniques based on gazetteers instead of on more sophisticated representation systems, is one of the main reasons because of which these techniques obtain worse results when applied to Biomedicine as opposed to other domains (Spasic et.al., 2005).

## 3. Techniques Applied to Biomedical Information Retrieval Systems

Information Retrieval systems apply Natural Language Processing tasks (BioNLP when applied to Biomedicine), such as the decomposition of a text into tokens, Part-Of-Speech-Tagging, noun phrase chunking and word sense disambiguation or coreference resolution. Tokenization must be performed differently in the biomedical domain, as it cannot be resolved straightforwardly by relying on white spaces and punctuation marks as explicit delimiters (e.g. [3H]R1881 is a single token). There are studies pointing out that POS-taggers adapted to the biomedical domain improve their effectiveness (Zhou et.al., 2004), and results have been compared in this matter (Clegg and Sheperd, 2005). There are tools adapted as well, such as the GENIA tagger, which analyzes English sentences and outputs the base forms, POS-tags, chunk tags and named entity tags. The GENIA tagger is trained not only on the Wall Street Journal corpus, but also on the GENIA and the PennBioIE corpora, so the tagger works well on various types of biomedical documents (Tsuruoka and Tsujii, 2004).

Another technique frequently used on information retrieval systems in general is Information Extraction. From 1998 on, there has been an increasing interest for the recognition of named entities in Biomedicine (BioNER), mainly for names of genes and genetic products, due to the Human Genome Project. Nowadays, BioNER is also applied to the recognition of AND, ARN, cell line, cell type, mutations, properties of the protein structures, etc.

BioNER systems (Table 4) have evolved similarly to the general purpose ones, going from techniques based on hand-made rules to systems based on supervised learning from tagged corpora. This is the approach used by most systems, although in BioNER the support of lexical resources is stronger, given the terminological problems present in the domain. These resources provide good rates in terms of precision, but not for recall, as it is virtually impossible to include every relevant entity in these lists because of the constant incorporation of new terms.

In return, semi-supervised techniques are less frequent in BioNER, although there are several works that use active learning and bootstrapping. It is interesting to see that the application of bootstrapping to NER appears in the late nineties and it is based on few initial examples, used as seeds in progressive learning to find new patterns capable of recognizing new entities. However, the use of these techniques in BioNER is more recent, and it is common to use much larger resources, such as dictionaries or ontologies, to tag corpora and create learning models (Morgan et.al., 2004) or to infer instances and lexical-semantic patterns for their capture (Xu et.al., 2009). These vast processes are justified by the variability of patterns, both in the own entities and in their context. The context may also prove confusing because of the semantic closeness of some categories of entities to capture (by composition for example, as is the case of genes and proteins). These factors make it difficult to progressively learn patterns based on a reduced number of tagged examples, so larger resources become essential for a reliable learning process.

The use of these techniques in the Web is also rare, even though it has been used, for example, to filter gene names, improving the F-score of the results by 0.17 (Dingare et.al., 2004). In any case, the tendencies in data mining are the same as in the Web (Baeza-Yates, 2009). The Web's semantic tagging, with methods both from the Semantic Web and from the Web 2.0, contributes with no doubt to the improvement of the results from IE techniques.

Table 4. Freely available BioNER tools

| Tool | Entities | Type | Main techniques |
|---|---|---|---|
| **ABNER** pages.cs.wisc.edu/~bsettles/abner/ | Proteins, DNA, RNA, cell line, cell type | Research | Supervised learning (on NLPBA and BioCreative). Trainable |
| **AbGene** ftp.ncbi.nlm.nih.gov/pub/tanabe/AbGene | Genes, proteins | Research | Based on statistically extracted rules (over MEDLINE abstracts) |
| **PIE** pie.snu.ac.kr | Protein, protein interactions | Research | NLP, based on dictionary- and supervised-learning |
| **BIORAT** bioinf.cs.ucl.ac.uk/downloads/biorat/ | Proteins, protein interactions | Research, Commercial (Ebisu) | Based on NLP, dictionary and predefined patterns with the GATE IE framework |
| **Lingpipe** alias-i.com/lingpipe/web/download.html | Genes, proteins and others | Commercial (Alias-i) | General IE tool based on supervised-learning (on Genia and MedPost). Trainable |

Finally, within the Information Extraction, another technique frequently applied to IR systems is the detection of relations. Tasks such as detecting functional properties of genes or interactions of proteins are gaining special relevance. In these tasks, the problems of BioNER extend to the multiple types of different relations we can find. For this reason, the support of Knowledge Organization Systems is here more noticeable than in other areas.

## 4. Information Retrieval Evaluation in the Biomedical Domain

The TREC conference used test collections from the biomedical domain for the evaluation of IR systems, but in the year 2000 there was already a track specific to Biomedicine. The test consisted on evaluating the capability of different systems for classifying OHSUMED documents (part of MEDLINE) with the MeSH categories. In TREC-2003 there was a retrieval track dedicated to Genomics, and in 2004 this track was centered on tagging genes and proteins in relevant documents. This way, it was attempted to emulate the manual process curators performed in the Mouse Genome Informatics, where they tagged genes with GO (Hersh, 2004). The last edition of this track took place in 2007, when the task consisted on responding questions containing entities whose type was defined within the question itself (e.g. "what [drugs] have been tested in mouse models of Alzheimer's disease?"). Another important forum in the area is BioCreative, which was carried out in 2004 and 2006, with the objective of recognizing entities and relations about genes and proteins. Later on, there were trials in other places, like ImageCLEF 2007 for retrieving medical images.

Precision rates for information retrieval and extraction tasks in the area are between 70 and 90 percent, while recall is around 70%. These figures are 15% lower than those achieved in other domains such as the journalistic (Ananiadou, 2006). Nevertheless, the rates achieved in the journalistic domain for NER, basis for the other tasks, are not better. Sometimes it has even been considered outperformed, with 90% success rates (Cunningham, 2005). However, it has been demonstrated that changing the sources used, even just in the document genre and not specifically in the domain, yield to remarkable loss of effectiveness (between 20 and 40 percent) (Poibeau and Kosseim, 2001). In the biomedical domain it has been observed that training with a tagged corpus and then evaluating with another one, leads to a 13% drop in the F-measure (Leser and Hakenberg, 2005). Since the tools are trained for a particular collection, their behavior for other collections is different, and it is expected to be different in real cases too. The terminological characteristics, such as the occurrence of many compound words and the necessity of knowledge from diverse sub-areas, make the tagging process quite difficult, with inter-annotator agreements between 75 and 90 percent for genes and proteins (Gaizauskas et.al., 2003). There is some work nowadays to improve the consistency between annotations, such as the development of a schema for semantic annotation in the public health domain (Kawazoe et.al., 2009).

Many works conclude that the evaluation of such retrieval systems in Biomedicine must be user-oriented, developing metrics and methods capable of measuring the user's satisfaction in real-life tasks (Leser and Hakenberg, 2005; Cohen and Hersh, 2005). To achieve that, there needs to be cooperation between the experts of the Information Retrieval and Extraction field, and the ones of the Biomedicine domain. Recent examples of this type of cooperation include the BioCreative 2004 workshop, and the TREC Genomics Track, both of which used assessments made by biological database curators in their normal workflow processes as the gold standard.

## 5. Conclusions

Biomedicine features many peculiarities as to the techniques and resources used for Information Retrieval. These features are very diverse, and they introduce many problems for IR systems, where the lacks of terminological consensus and of patterns in the terminology used are two of the most important ones. The former affects the construction and integration of Knowledge Organization Systems and their application to IR systems, while the latter limits the use of machine-learning techniques. In the face of the traditional consensus problems in Biology, regarding both nomenclature and organization, there prevails the use of the Internet and the standardization of resources, with formats and construction rules.

Initiatives for normalizing terminology, such as the one carried out by the Human Genome Organization (HUGO, http://www.hugo-international.org) and standard tagging methodologies, can contribute to the improvement of the retrieval success rates, as well as the effective use of semantic tagging tools to assist in corpora tagging process. This process is a way to achieve a constant updating of the biomedical resources, which provides support for the information retrieval systems.